\documentclass[letterpaper, 10 pt, conference]{ieeeconf}  
\usepackage{graphicx}
\usepackage{amsmath}
\usepackage{mathrsfs}
\usepackage[style=ieee]{biblatex}

\usepackage{xcolor}
\usepackage{balance}
\DeclareSourcemap{
  \maps{
    \map{
      \pertype{article}
      \step[fieldset=url, null]
      \step[fieldset=doi, null]
      \step[fieldset=issn, null]
      \step[fieldset=isbn, null]
      \step[fieldset=note, null]
      \step[fieldset=editor, null]
      \step[fieldset=urldate, null]
      \step[fieldset=file, null]
    }
  }
}
\DeclareSourcemap{
  \maps{
    \map{
      \pertype{inproceedings}
      \step[fieldset=url, null]
      \step[fieldset=doi, null]
      \step[fieldset=issn, null]
      \step[fieldset=isbn, null]
      \step[fieldset=note, null]
      \step[fieldset=editor, null]
      \step[fieldset=urldate, null]
      \step[fieldset=file, null]
    }
  }
}
\DeclareSourcemap{
  \maps{
    \map{
      \pertype{incollection}
      \step[fieldset=url, null]
      \step[fieldset=doi, null]
      \step[fieldset=issn, null]
      \step[fieldset=isbn, null]
      \step[fieldset=note, null]
      \step[fieldset=editor, null]
      \step[fieldset=urldate, null]
      \step[fieldset=file, null]
    }
  }
}
\DeclareSourcemap{
  \maps{
    \map{
      \pertype{misc}
      \step[fieldset=url, null]
      \step[fieldset=doi, null]
      \step[fieldset=issn, null]
      \step[fieldset=isbn, null]
      \step[fieldset=note, null]
      \step[fieldset=editor, null]
      \step[fieldset=urldate, null]
      \step[fieldset=file, null]
    }
  }
}

\IEEEoverridecommandlockouts                              

\title{\LARGE \bf
Reduced-Order Model-Based Gait Generation\\for Snake Robot Locomotion using NMPC
}

\author{Adarsh Salagame$^{1}$, Eric Sihite$^{2}$, Milad Ramezani$^{3}$, Alireza Ramezani$^{1*}$
\thanks{$^{1}$This author is with the Department of Electrical and Computer Engineering, Northeastern University, Boston MA
        {\tt\small salagame.a, a.ramezani@northeastern.edu*}}%
\thanks{$^{2}$ This author is with California Institute of Technology, Pasadena CA
		{\tt\small esihite@caltech.edu}}%
\thanks{$^{3}$Author is with CSIRO Robotics, DATA61, CSIRO, Brisbane, Australia. Email: 
        {\tt\small milad.ramezani@data61.csiro.au}}%
\thanks{$*$Indicates the corresponding author.}
}

\begin{document}

\maketitle
\thispagestyle{empty}
\pagestyle{empty}

\begin{abstract}

This paper presents an optimization-based motion planning methodology for snake robots operating in constrained environments. By using a reduced-order model, the proposed approach simplifies the planning process, enabling the optimizer to autonomously generate gaits while constraining the robot’s footprint within tight spaces. The method is validated through high-fidelity simulations that accurately model contact dynamics and the robot’s motion. Key locomotion strategies are identified and further demonstrated through hardware experiments, including successful navigation through narrow corridors. 

\end{abstract}

\section{INTRODUCTION}

Optimization-driven path planning and control strategies 
\cite{ding_real-time_2019,carius_trajectory_2018,droge_optimal_2012,chang_optimal_2018,hicks_method_2005,mohammadi_optimal_2009} 
have become pivotal methodologies for managing diverse, contact-intensive systems in real-world experimental settings. These approaches have been widely applied across various locomotion and manipulation modalities, including dynamic walkers, crawlers (slithering locomotion), and manipulators. Their efficacy is particularly notable in rapid contact planning, especially in terrestrial environments.

Among these systems, crawlers present particularly complex, contact-rich challenges. Point-contact models 
\cite{katayama_model_2023, han_3d-slip_2022}, 
commonly used for legged robots, are generally more compatible with optimization techniques compared to systems involving extensive contact interactions, such as snake robots. However, these models do not fully capture the surface contact dynamics that govern crawling systems' movement (e.g., crawling and burrowing through dirt, sand, or other viscous environments). Despite these limitations, point-contact models remain valuable, especially in scenarios such as crawling on flat, rigid surfaces.

\begin{figure}
    \centering
    \includegraphics[width=0.9\linewidth]{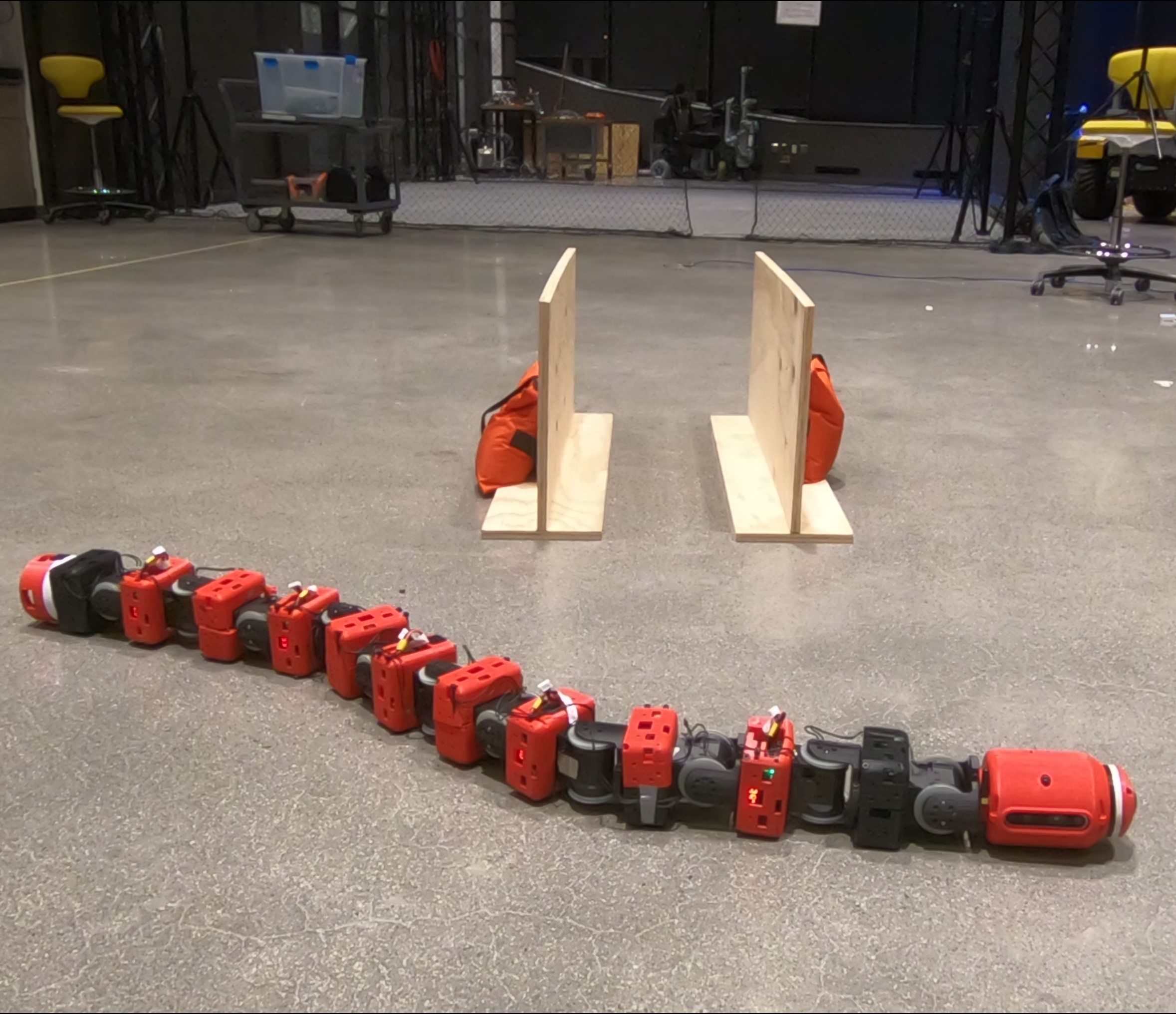}
    \caption{Shows COBRA \cite{salagame_loco-manipulation_2024,jiang_snake_2024,jiang_hierarchical_2024, salagame_heading_2024,salagame_how_2024} traversing through a narrow path.}
    \label{fig:cover-image}
    \vspace{-5mm}
\end{figure}

Given the intricate dynamics inherent in slithering systems, which encompass sophisticated contact dynamics 
, there arises a pressing need for enhanced modeling and control methodologies. These tools are indispensable for orchestrating body movements through the modulation of joint torques, ground reaction forces, and the coordination of contact sequences comprising timing and spatial positioning. Snake locomotion embodies a spectrum of primitives to diverse environments and challenges:

\begin{itemize}
    \item Lateral undulation, as exemplified in studies by \cite{wiriyacharoensunthorn_analysis_2002, ma_analysis_2003, ma_analysis_1999}, relies on anisotropic friction to propel snakes forward in a sinusoidal trajectory. 
    \item Rectilinear motion, presented in works such as \cite{rincon_ver-vite_2003, ohno_design_2001}, involves the controlled compression and expansion of scales to facilitate longitudinal movement, ideal for maneuvering through constrained spaces. 
    \item The sidewinding gait, as demonstrated in \cite{liljeback_modular_2005, burdick_sidewinding_1994}, is employed on slippery or sandy terrains, featuring a sinusoidal motion for lateral displacement. 
    \item In confined settings, snakes adopt the concertina gait, outlined in \cite{shan_design_1993}, characterized by coiling and uncoiling actions to progress longitudinally.
    \item Unconventional gaits such as the inchworm, slinky, lateral rolling, and tumbling locomotion, proposed in studies like \cite{yim_new_1994, rincon_ver-vite_2003}, exploit the articulated structure of the snake's body to manifest unique locomotive patterns.
\end{itemize}

\begin{figure}
    \centering
    \includegraphics[width=0.9\linewidth]{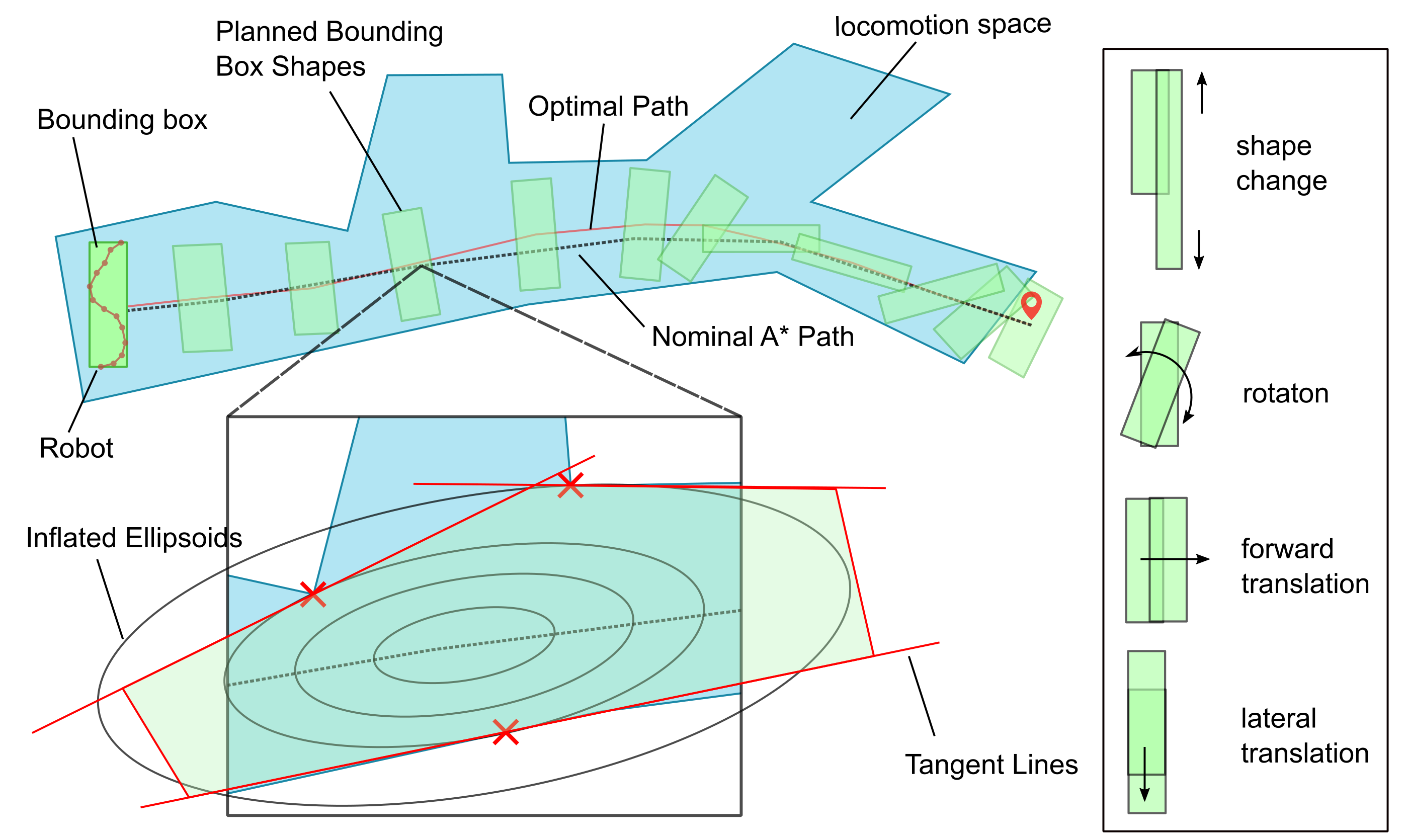}
    \caption{Highlights our approach to formulating COBRA's path planning and tracking problem by reducing its multi-DOF posture manipulation to a few key actions, including shape adjustment, forward and lateral translation, and rotation, all within a bounding box that encapsulates COBRA.}
    \label{fig:problem-statement}
    \vspace{-3mm}
\end{figure}

So far, we have seen a rich body of control works including concepts based on backbone curve method \cite{huang_unified_2024}, floating frame of reference \cite{melo_modular_2015}, virtual chassis \cite{rollinson_virtual_2012}, torsion and curvature used to describe snake shape \cite{melo_modular_2015}, \cite{nonhoff_economic_2019} economic MPC, MPC for viscous environment \cite{hannigan_automatic_2020}, and Frenet-Serret framework \cite{yamada_study_2006} 
. It can be seen that while the predominant focus of snake robotics research lies in emulating snake locomotion and replicating its distinctive movement patterns using central pattern generators, contact-implicit optimization based on reduced-order models remains underutilized for obvious reasons (e.g., failure to capture burrowing in viscous environments). 

Leveraging optimization techniques \cite{wanner_model_2022,salagame_non-impulsive_2024,salagame_dynamic_2024,sihite_optimization-free_2021,mandralis_minimum_2023,salagame_quadrupedal_2024,krishnamurthy_narrow-path_2024,krishnamurthy_thruster-assisted_2024} on reduced-order models of snakes that use simplified (i.e., multiple point-contact) rigid contact models, although not fully capturing the complexity of a snake's contact-rich locomotion, can enable the control of these simplified models in most practical scenarios. Consequently, it allows for the control of the full dynamics they represent through optimal joint movements that respect complementarity conditions. These frameworks can leverage the inherent redundancy and intricate articulation of snake bodies for acyclic locomotion and manipulation (loco-manipulation), enabling fast path planning and replanning.

In this paper, we specifically examine path planning and tracking over flat, rigid surfaces with randomly scattered obstacles. The control design for this type of locomotion involves tracking the position, orientation, and shape of the snake. If these obstacles dynamically change positions, rapid path planning and tracking become essential.

\section{Problem Statement}


Figure~\ref{fig:problem-statement} shows our view towards path planning and tracking in snake-type locomotion based on reduced-order modeling.

Several studies in the literature tackle the problem of automatic gait generation for snake robots, most focus on movement along the robot’s length, relying on sinusoidal multi-joint body motions and anisotropic friction with the ground to generate forward velocity \cite{ouyang_motion_2018, nor_cpg-based_2014}.

Some snake-like robots have isotropic friction contacts and use gaits that have intermittent ground interactions and 3D posture manipulation to generate movement \cite{huang_unified_2024, wang_cpg-inspired_2017}. Due to the large number of discontinuous ground interactions across the body of the robot, dynamic modeling is challenging, and closed loop locomotion control of such snake robots is an ongoing area of research. 

We use a reduced-order model to abstract the robot's shape and movement, avoiding internal oscillations and ground contact modeling. This model represents the snake robot’s extremities as a bounding box for path planning in constrained spaces (Fig. \ref{fig:problem-statement}). A high-level planner generates a nominal CoM trajectory via A*, then optimizes the path and bounding box shape for efficient locomotion. Following \cite{liu_planning_2017}, it inflates ellipsoids along the path to define a convex safety corridor. Tangents at obstacle contacts form corridor boundaries (Fig. \ref{fig:problem-statement}). Using these boundaries as constraints, we can implement a gait generator that plans optimal trajectories and bounding box shapes to reach the desired goal.

In this work, we assume knowledge of the safety corridor and emphasize the reduced-order modeling framework and the development of a gait generation using Nonlinear Model Predictive Control (NMPC). We use a Central Pattern Generator (CPG) for low level joint control, and use the NMPC to drive the parameters of the CPG to achieve gait generation. Fig. \ref{fig:problem-statement} shows different locomotion modes each created by changing CPG parameters that allow movement through spaces of different sizes. As a proof of concept, we demonstrate the effectiveness of this approach by using CPG to control our snake robot COBRA to travel through a narrow corridor.


This paper is organized as follows. Section \ref{sec:rom} details the development of the reduced order model for COBRA. In Section \ref{sec:low} we discuss low-level joint control using CPG, followed by a kinematics-based motion model in Section \ref{sec:motion}. In Section \ref{sec:nlp} we use this motion model in an NMPC-based control scheme for autonomous gait generation. Finally, we present experimental and simulation results evaluating components of the presented framework and demonstrating the effectiveness of CPG based locomotion modes in navigating narrow paths using predetermined parameters.

\section{Reduced-Order Modeling}
\label{sec:rom}

\begin{figure}
    \centering
    \includegraphics[width=0.9\linewidth]{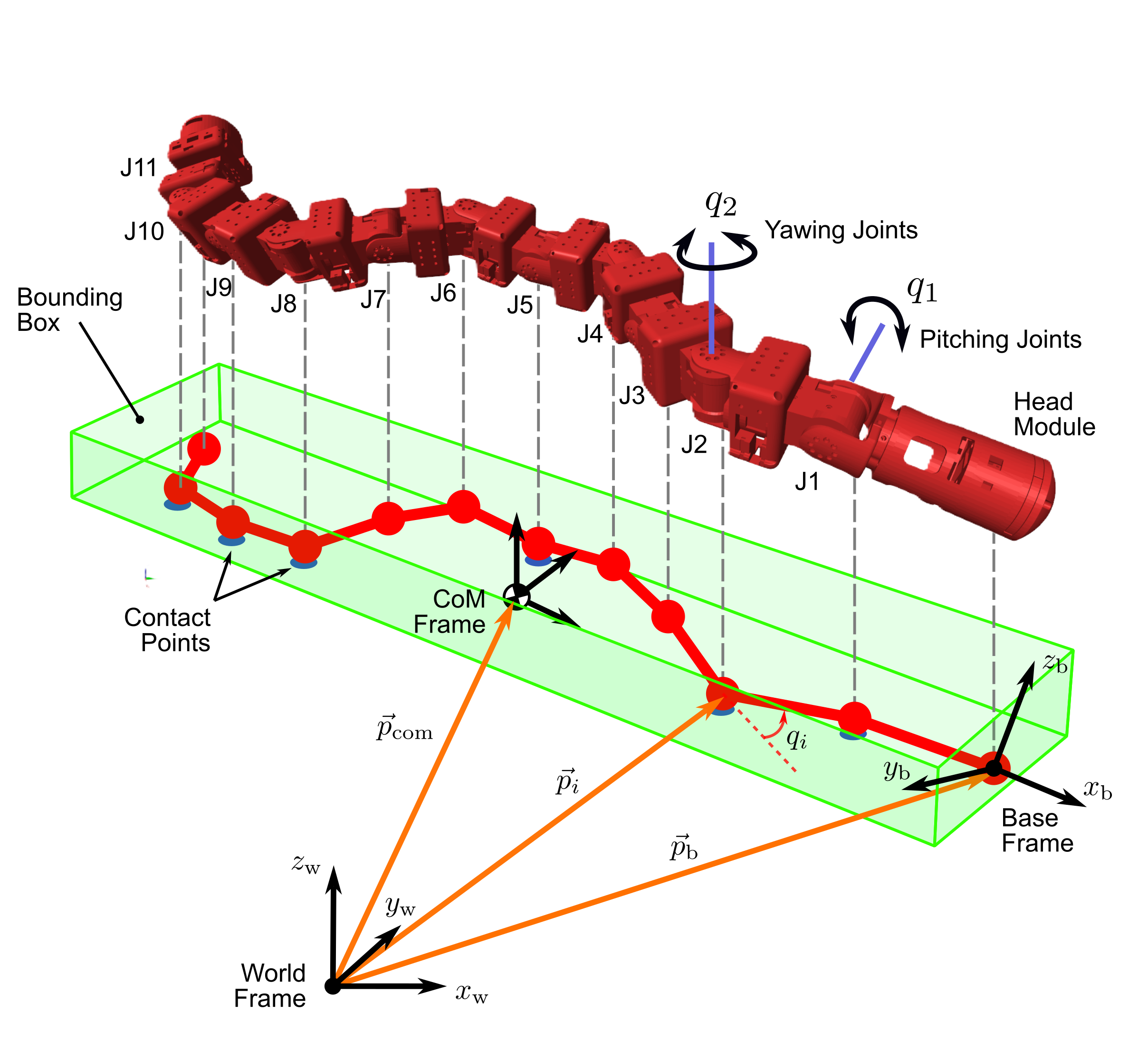}
    \caption{Shows the Reduced-Order Model for our snake robot COBRA}
    \label{fig:rom}
    \vspace{-3mm}
\end{figure}

Figure~\ref{fig:rom} shows the overall process of generating our reduced-order model. To derive the Reduced-Order Model (ROM), each link of the snake robot is modeled as a point mass positioned at the link's origin. Fig. \ref{fig:rom} illustrates the robot's kinematic structure, starting with the base frame located at the head module, followed by 12 links and 11 joints. Odd joints numbered $J_1 \dots J_{11}$ from the head module are pitching joints and even joints numbered $J_2 \dots J_{10}$ are yawing joints. Each joint consists of a single degree of freedom position controlled servo actuator powered independently and controlled from the head module.

The robot state taking the base frame as reference is defined by the vector $\mathbf{X}_b = \begin{bmatrix}p_\text{b} & \Phi_b & q\end{bmatrix}^\top$, where $\vec p_\text{b}$ represents the position of the base frame, $\Phi_b$ describes its orientation using XYZ-Euler angles, and $\vec q$ denotes the joint angles of each link. This state vector is fully observable through onboard sensors, including a stereo camera, an Inertial Measurement Unit (IMU), and absolute joint encoders.

Using forward kinematics, the position of each link in the world frame, $\vec p_i$, is computed. The resulting $3 \times N$ matrix given by
\begin{equation}
    \mathbf{P}_\text{links} = \mathscr{F}(\mathbf{X}_b)
    \label{eq:fk}
\end{equation} 
encapsulates the positions of all links in the world frame, where $N$ is the number of links in the robot.

\subsection{Floating Base Frame for ROM}

Snake robot gaits typically involve oscillatory motions across all links. To describe the motion of the entire robot in a way that minimizes the influence of these oscillations, a floating reference frame is required—one that remains mostly stable and independent of the individual link movements. Following the approach in \cite{rollinson_virtual_2012}, we define a center of mass (COM) frame as this floating reference frame. The position of the COM in world coordinates is calculated as:
\begin{equation}
    \mathbf{P}_\text{com} = \frac{\sum_{i=1}^{N} m_i \vec{p}_i}{\sum_{i=1}^{N} m_i}
\end{equation}

Where $m_i$ is the mass of the $i^\text{th}$ link. We stack all $\vec{p}_i$ link positions and form a wide matrix called $\mathbf{P}_\text{links}$. Then, we form another wide matrix $\tilde{\mathbf{P}}_\text{links}$ where the entries are the relative position of the links with respect to the snake COM $\vec p_\text{com}$ given by
\begin{equation}
    \tilde{\mathbf{P}}_\text{links} = \mathbf{P}_\text{links} -\left[\mathbf{P}^\top_\text{com},\dots,\mathbf{P}^\top_\text{com}\right]
\end{equation}

The orientation of the center of mass frame is determined using Singular Value Decomposition (SVD):
\begin{equation}
    \tilde{\mathbf{P}}_\text{links} = \mathbf{U}\Sigma \mathbf{V}^\top
\end{equation}
Here, $\mathbf{U} = \begin{bmatrix}\hat x & \hat y & \hat z\end{bmatrix}$ provides the orthogonal basis vectors of the row space of $\tilde{\mathbf{P}}_\text{links}$. However, without any constraints, the axes derived from $\mathbf{U}$ may not adhere to the right-hand rule, nor do they guarantee stable orientation. To ensure consistency with previous time steps, the directions of $\hat x$ and $\hat z$ are flipped if necessary to align with the previous orientation, and the cross product is used to compute $\hat y$. This process produces the rotation matrix $\mathbf{R}_\text{com}$ that defines the orientation of the center of mass frame with respect to the world.

\subsection{Bounding Box Definition of ROM}

To further simplify the reduced-order model, we define a bounding box that encapsulates the snake robot's footprint in 3D space at any given time. This bounding box abstracts the robot's internal configuration, focusing instead on its external spatial extent. The position of each link in the center of mass frame is computed as:
\begin{equation}
    \mathbf{\bar P}_\text{links} = \mathbf{R}_\text{com}~\tilde{\mathbf{P}}_\text{links}
\end{equation}
From these positions, the bounding box is determined by calculating the extents along each axis:
 \begin{equation}
      \Delta_k  = \frac{1}{2}\left[\text{max}(\mathbf{\bar P}_\text{links} \hat k) - \text{min}(\mathbf{\bar P}_\text{links} \hat k)\right]  \\
 \end{equation}
$\text{where}~k \in \begin{bmatrix}x & y & z\end{bmatrix}$. This yields the dimensions of the bounding box along the $\hat x$, $\hat y$ and $\hat z$ axes, effectively representing the robot's overall footprint at each time step.

\subsection{Contact Signal Estimation}
Assuming a flat ground surface and quasi-static robot movements, the contact signal for the reduced-order model can be estimated using an $N \times 1$ vector $\mathbf{C}$. This vector identifies which links are in contact with the ground by checking if each link is within a threshold $\varepsilon$ from the base of the bounding box. Mathematically, this is expressed as:
\begin{equation}
    \mathbf{C} = \begin{cases}
        0, \quad \mathbf{\bar P}_\text{links} \hat z > \vec p_\text{com} - (\Delta_z - \varepsilon) \\
        1, \quad \mathbf{\bar P}_\text{links} \hat z \leq \vec p_\text{com} - (\Delta_z - \varepsilon)
    \end{cases}
    \label{eq:contact}
\end{equation}

All components of the reduced-order model, including the center of mass frame, bounding box, and contact signal are illustrated in Figure \ref{fig:rom}. With the above framework, we establish a mapping from the observable robot base state $X_\text{b}$ to the reduced-order model state $\{\mathbf{P}_\text{com},~\mathbf{R}_\text{com},~ \mathbf{C}\}$.

\section{Low-level Joint Control}
\label{sec:low}
\begin{figure}
    \centering
    \includegraphics[width=0.7\linewidth]{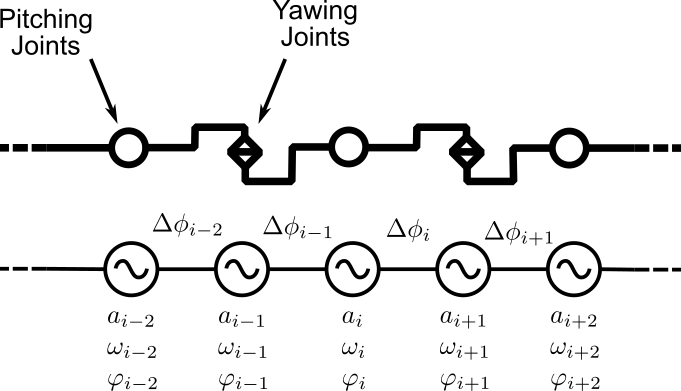}
    \caption{Depicts kinematic configuration and CPG parameters}
    \label{fig:cpg}
    \vspace{-5mm}
\end{figure}

COBRA, like many snake robots, features alternating single-degree-of-freedom pitching and yawing joints, as shown in Figure \ref{fig:cpg}. Low-level control of these joints is achieved through a Central Pattern Generator (CPG), which generates sinusoidal signals for each joint. The parameters governing the CPG are defined by the vector $[a,~\omega,~\varphi]^\top$, representing the amplitude, frequency, and phase of each oscillator, respectively. To ensure smooth, continuous joint movements, a CPG model incorporating coupling between adjacent joints, based on the work of \cite{bing_cpg-based_2017}, is employed. The relative phase between consecutive joints is defined as $\phi_i = \varphi_{i+1} - \varphi_i$. Using this, for internal CPG state $\mathbf{X}_\text{cpg} = [\theta,~r,~\dot r]^\top$ representing the instantaneous phase, amplitude and rate of change amplitude of each oscillator, the internal state update model is described as:  
\begin{equation*}
    \begin{bmatrix}
        \dot \theta \\ 
        \dot r \\ 
        \ddot r
    \end{bmatrix} = 
    \begin{bmatrix}
        \mu \mathbf{A} & 0 & 0 \\
        0 & 0 & 1 \\
        0 & -\gamma^2 & -\gamma
    \end{bmatrix}
    \begin{bmatrix}
        \theta \\
        r \\
        \dot r
    \end{bmatrix} + 
    \begin{bmatrix}
        0 & 1 & -\mu \mathbf{B} \\
        0 & 0 & 0 \\
        \gamma^2 & 0 & 0
    \end{bmatrix}
    \begin{bmatrix}
        a \\
        \omega \\
        \phi
    \end{bmatrix}
\end{equation*}

Here, $\gamma$ and $\mu$ are parameters regulating the intensity of tracking the desired phase and amplitude, while $\mathbf{A}_{11 \times 11}$ and $\mathbf{B}_{11 \times 10}$ are are the coupling matrices for phase and amplitude, defined as:


$$
\begin{array}{cc}
     A = \begin{bmatrix}
  1 & 1  &   &  \\
  1 & -2 &   \ddots &  \\
    & \ddots   &  \ddots &  \\
    &    &   &  -2 & 1 \\
    &    &   & 1 & 1
\end{bmatrix} &

B = \begin{bmatrix}
 1     &    &    \\
-1     & \ddots  & \\
       &  \ddots  &  1\\
       &    & -1
\end{bmatrix}
\end{array}
$$


The desired joint angles generated by the CPG are then given by:
\begin{equation}
    \begin{bmatrix}
        q\\
        \dot q\\
        \ddot q
    \end{bmatrix} = 
    \begin{bmatrix}
        r \text{sin}(\theta) \\
        \dot r \text{sin}(\theta) + r \dot \theta \text{cos}(\theta) \\
        \ddot r \text{sin}(\theta) + 2 \dot r \dot \theta \text{cos}(\theta) + r \ddot \theta \text{cos}(\theta) - r \dot \theta^2 \text{sin}(\theta)
    \end{bmatrix}
    \label{eq:cpg}
\end{equation}

\section{Kinematic Motion Model}
\label{sec:motion}
From the forward kinematics equation (\ref{eq:fk}), the velocities and accelerations of each link in the world frame can be computed using the Jacobian matrix as follows:

\begin{equation}
    \begin{bmatrix}
        \mathbf{\dot P}_\text{links} \\
        \mathbf{\ddot P}_\text{links}
    \end{bmatrix} = \mathbf{J} \begin{bmatrix}
        \dot X_\text{b} \\
        \ddot X_\text{b}
    \end{bmatrix}
\end{equation}

Here, $\mathbf{J}$ is the Jacobian matrix of the mapping $\mathscr{F}$ with respect to $X_\text{b}$ and its time derivatives $\dot X_\text{b}$. Under the assumption of quasi-static movements, we impose no-slip constraints on all links in contact with the ground by setting the $x$ and $y$ components of velocity and acceleration to zero:

\begin{equation}
    \begin{array}{cc}
        \begin{bmatrix} \hat x & \hat y & 0\end{bmatrix}~ \mathbf{\dot P}_\text{links}~ \text{diag}(\mathbf{C})  &= 0\\ 
        \begin{bmatrix} \hat x & \hat y & 0\end{bmatrix}~ \mathbf{\ddot P}_\text{links}~ \text{diag}(\mathbf{C})  &= 0
    \end{array}
    \label{eq:constraint}
\end{equation}

From equation \ref{eq:cpg}, the values of $\{\dot q,~\ddot q\}$ are known, leaving $12$ unknown states in equation \ref{eq:constraint}:
$$[\dot p_\text{b},~ \dot\Phi_\text{b},~ \ddot p_\text{b},~\ddot \Phi_\text{b}] \in\mathcal{R}^3$$

This results in an over-determined system, which can be solved using a least-squares approach to obtain $\ddot X_\text{b}$ for the current time step. This is numerically integrated to obtain $X_\text{b}$ for the next time step, which is mapped with the reduced-order model state based on the framework presented in Section \ref{sec:rom}.

With this, we complete a modeling framework that takes in the current observable state $\mathbf{X}_\text{b}$ and CPG parameter inputs $\begin{bmatrix}a & \omega & \varphi\end{bmatrix}^\top$ and predicts the motion and bounds of the bounding box. In the following section, we present a NLP-based tracking controller that generates CPG parameter inputs to achieve position trajectory tracking.
\label{sec:nlp}
\begin{figure}
    \centering
    \includegraphics[width=0.9\linewidth]{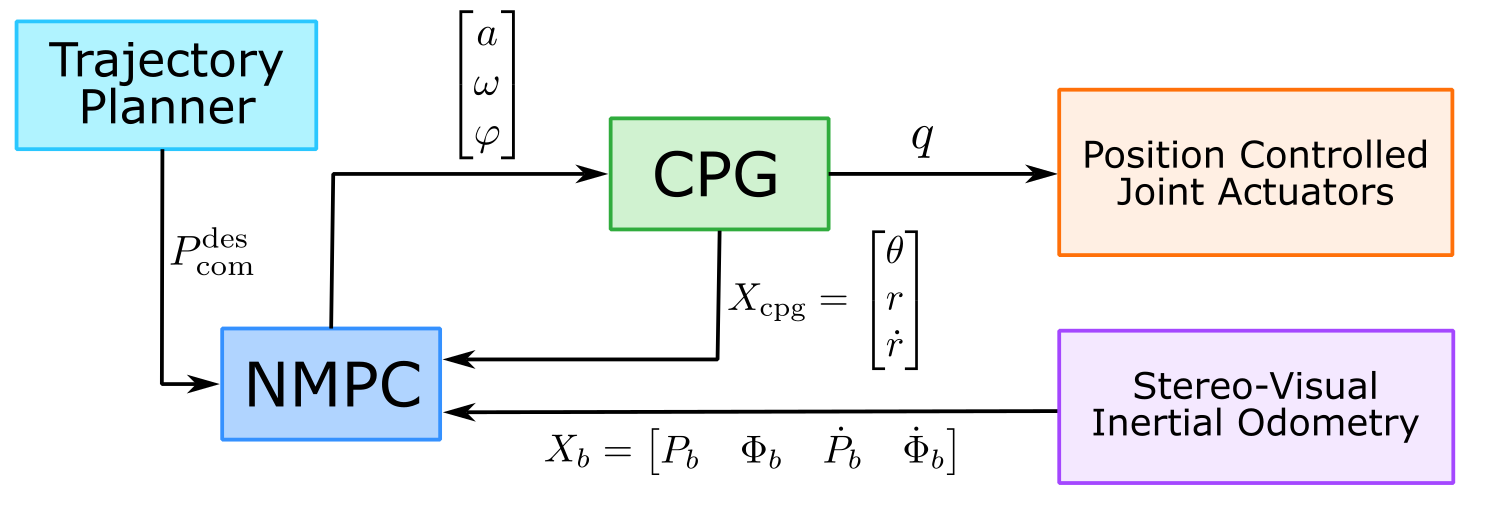}
    \caption{Shows an overview of the system architecture}
    \label{fig:system-overview}
    \vspace{-5mm}
\end{figure}
\section{Nonlinear Model Predictive Control (NMPC)}
We now formulate an NMPC problem that outputs CPG parameters for closed-loop gait generation in constrained environments. Figure \ref{fig:system-overview} shows the full proposed system architecture. The NMPC takes the desired goal position for the center of mass (CoM), denoted as $P_\text{com}^\text{des}$ as the reference for gait generation.

The decision vector for the optimization process, denoted as $\mathbf{X}$ consists of the CPG parameters applied over the entire horizon as well as the CoM positions at each step. The decision vector is expressed as

$$
\mathbf{X} = \begin{bmatrix}a & \omega & \varphi & \mathbf{P}_\text{com,1} & \dots & \mathbf{P}_\text{com,N}\end{bmatrix}^\top
$$

where $P_\text{com,i}$ represents the CoM position at each time step $t_i$ within the prediction horizon. The decision vector is constrained by bounds on both the CPG parameters and the CoM positions, denoted by $\mathbf{X}_\text{min}, ~\mathbf{X}_\text{max}$. Additionally, the size of the bounding box $[\mathbf{P}_\text{com}-\Delta, \mathbf{P}_\text{com}+\Delta]$ is constrained to be within the safe corridor. The objective of the optimization problem is to minimize the distance between the desired CoM position and the actual CoM position, represented as
\vspace{-2mm}
$$
\mathcal{J}(\mathbf{X)} = \sum_{i=1}^N \lvert\lvert \mathbf{P}_\text{com,i} - \mathbf{P}_\text{com}^\text{des} \rvert\rvert^2
$$

The predicted CoM position at each step, $P_\text{com}^\text{pred}(t_i)$, is determined by the motion model presented in Section \ref{sec:motion}. This prediction is a function of the CPG parameters set by the optimizer for the horizon. To compute this prediction, the NMPC utilizes the robot's state and the internal state of the CPG at the start of the horizon, advancing the model forward. The robot state is estimated using stereo visual-inertial odometry with an Intel RealSense D435i camera that is mounted in COBRA's head. The detailed implementation of state estimation will be discussed in a future publication. The predicted CoM positions is constrained to be equal to the CoM position set by the optimizer over the decision horizon $P_\text{com}(t_i)$. The full formulation of the NLP problem can then be written as 

$$
\begin{array}{cc}
\begin{array}{c}
\text{minimize:} \\
\mathbf{X}
\end{array}
&
\mathcal{J}(\mathbf{X)} = \sum_{i=1}^N \lvert\lvert \mathbf{P}_\text{com,i} - \mathbf{P}_\text{com}^\text{des} \rvert\rvert^2
\\
\begin{array}{c}
\text{subject to:}
\end{array}
&
\begin{array}{rcl}
  \mathbf{X}_\text{min} \le &  \mathbf{X}   & \le \mathbf{X}_\text{max} \\
  0 \le &\mathbf{P}_\text{com}^\text{pred} - \mathbf{P}_\text{com}& \le 0 \\
  p_\text{min} \le & [\mathbf{P}_\text{com} \pm \Delta] & \le p_\text{max}
\end{array}
\end{array}
$$

where $(p_\text{min},~p_\text{max})$ are the boundaries of the safety corridor in 2D. 

\section{Results}
\label{sec:results}
\begin{figure}
    \centering
    \includegraphics[width=0.9\linewidth]{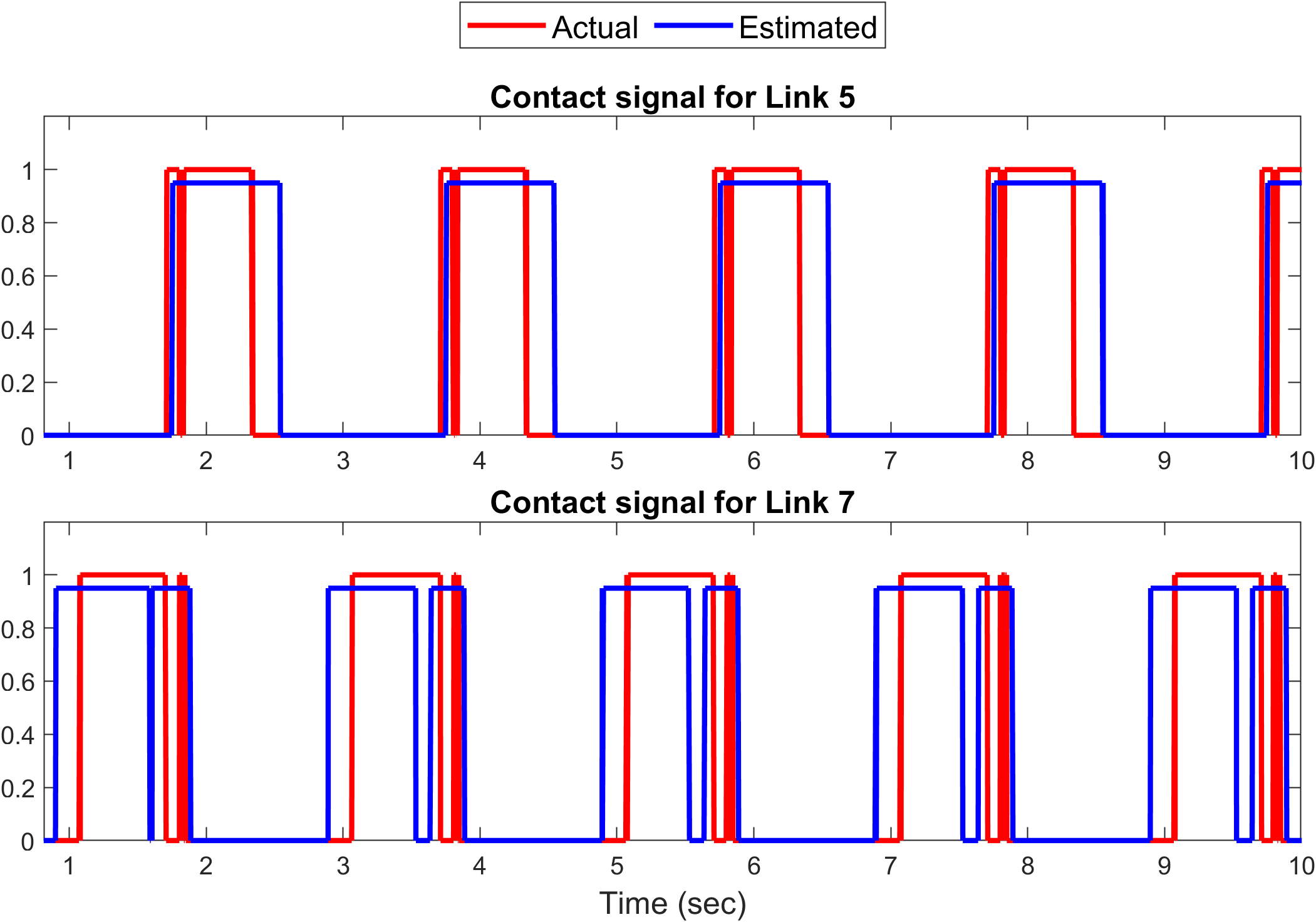}
    \caption{Comparison of the contact signal estimate from equation \ref{eq:contact} to ground truth from the simulation for two links with the ground truth simulation data for two links. The signals are binary, but a slight offset has been applied to the y-axis to facilitate clearer visualization.}
    \label{fig:contact-signal}
    
\end{figure}

\begin{figure}
    \centering
    \includegraphics[width=0.9\linewidth]{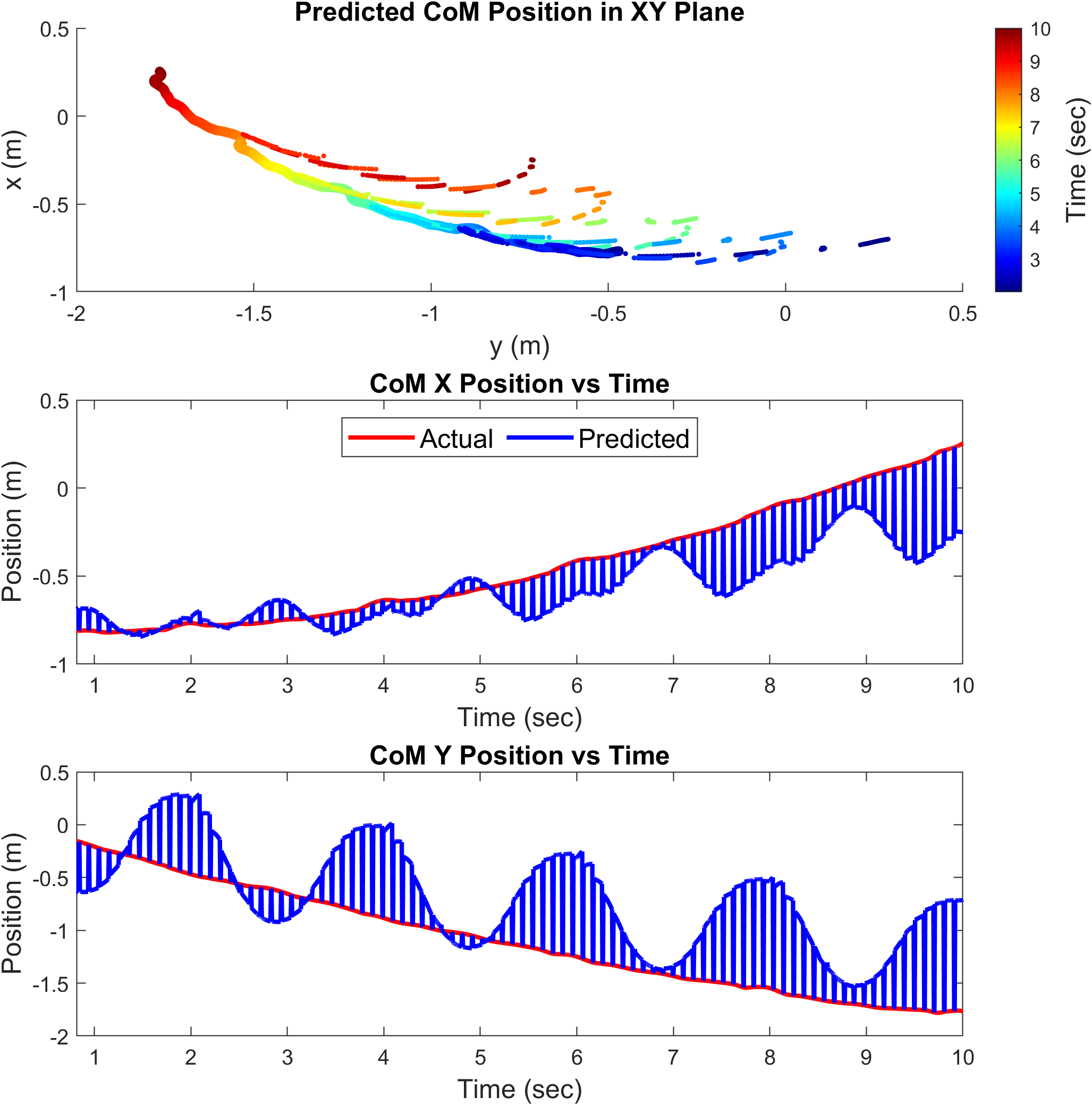}
    \caption{(Top) Shows prediction of CoM position in XY-Plane with color indicating time along the trajectory. The thick colored line represents the actual trajectory, while the thinner colorized offshoots represent predictions at each step. (Below) X and Y position predictions of the CoM as a function of time}
    \label{fig:com-pred}
    \vspace{-10mm}
\end{figure}
\begin{figure}
    \centering
    \includegraphics[width=0.9\linewidth]{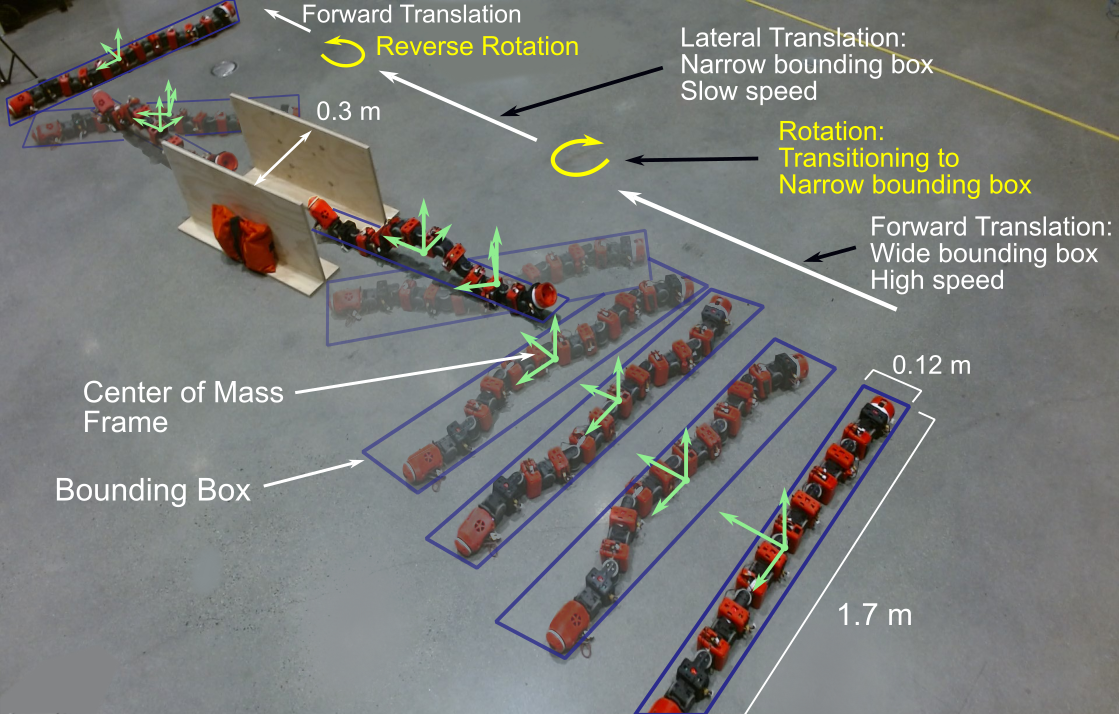}
    \caption{Shows experimental results with COBRA executing CPG gaits with predetermined parameters to use the identified key actions to travel through a narrow corridor. Also shown is the CoM positions and bounding box for each snapshot.}
    \label{fig:experiment}
    \vspace{-2mm}
\end{figure}
\begin{figure}
    \centering
    \includegraphics[width=0.9\linewidth]{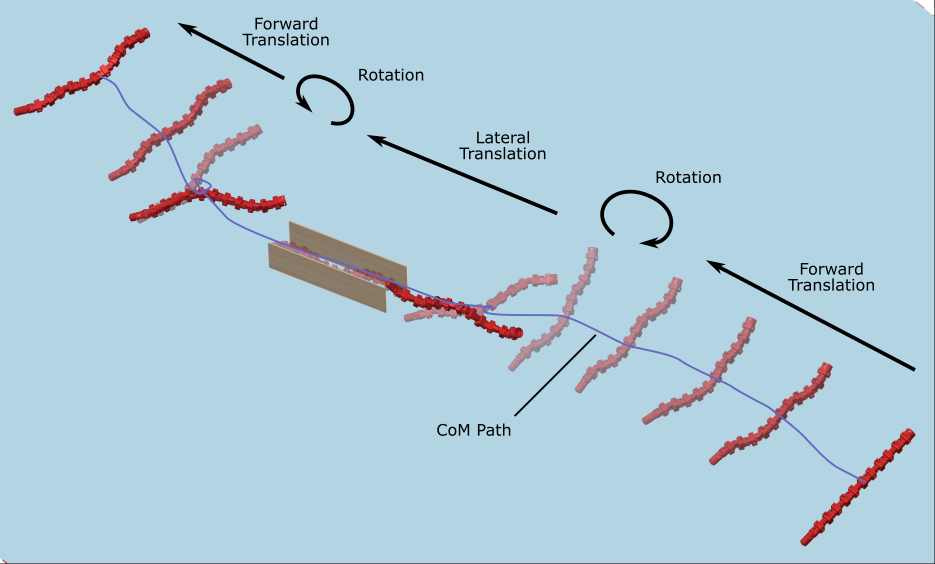}
    \caption{Demonstrates narrow path locomotion in simulation with the same CPG generated gaits used on hardware experiment}
    \label{fig:sim}
    \vspace{-3mm}
\end{figure}
\begin{figure}
    \centering
    \includegraphics[width=0.8\linewidth]{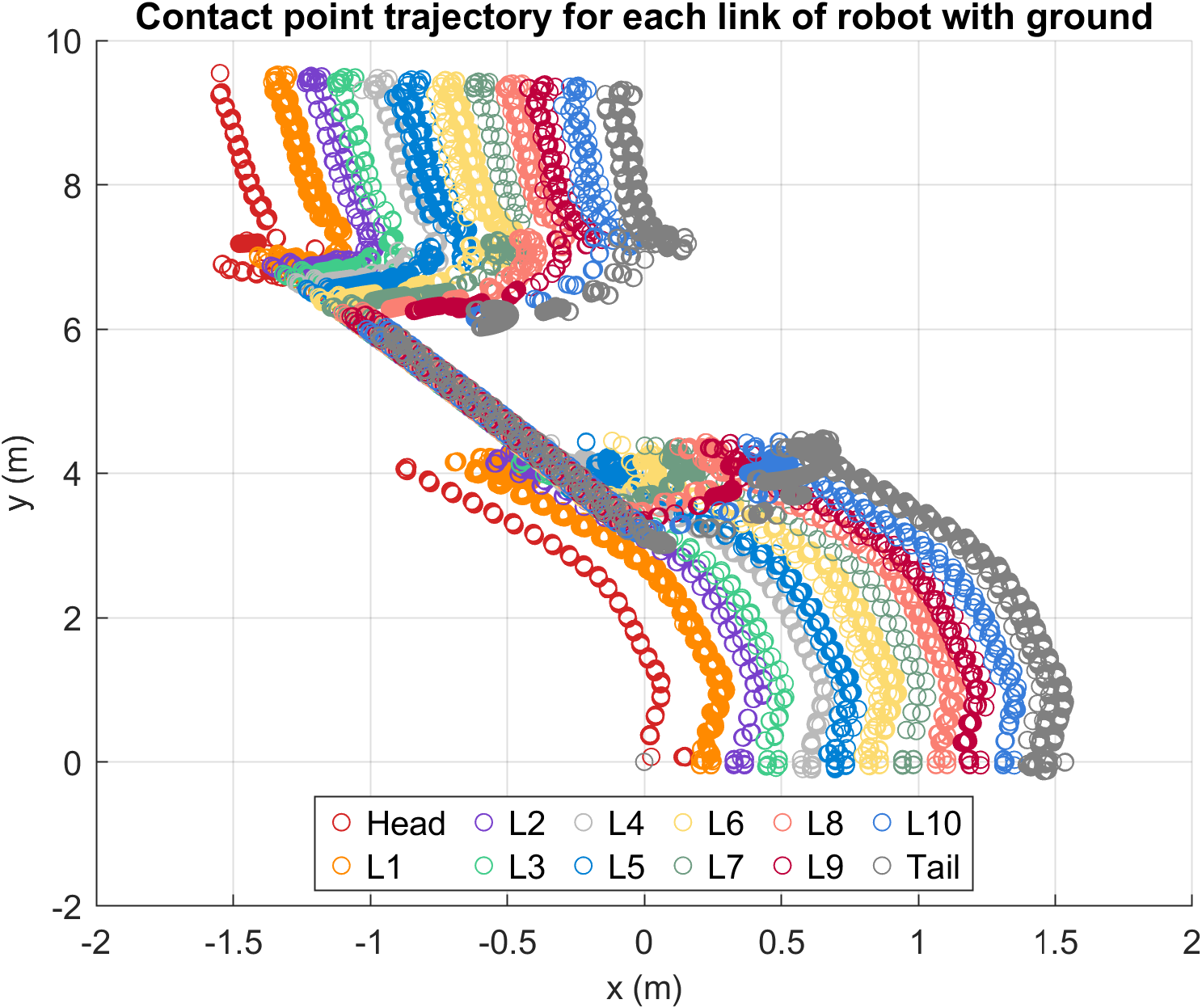} 
    \caption{Shows ground contact locations for each link throughout the locomotion in simulation}
    \label{fig:contact}
\end{figure}
\begin{figure}
    \centering
    \includegraphics[width=0.8\linewidth]{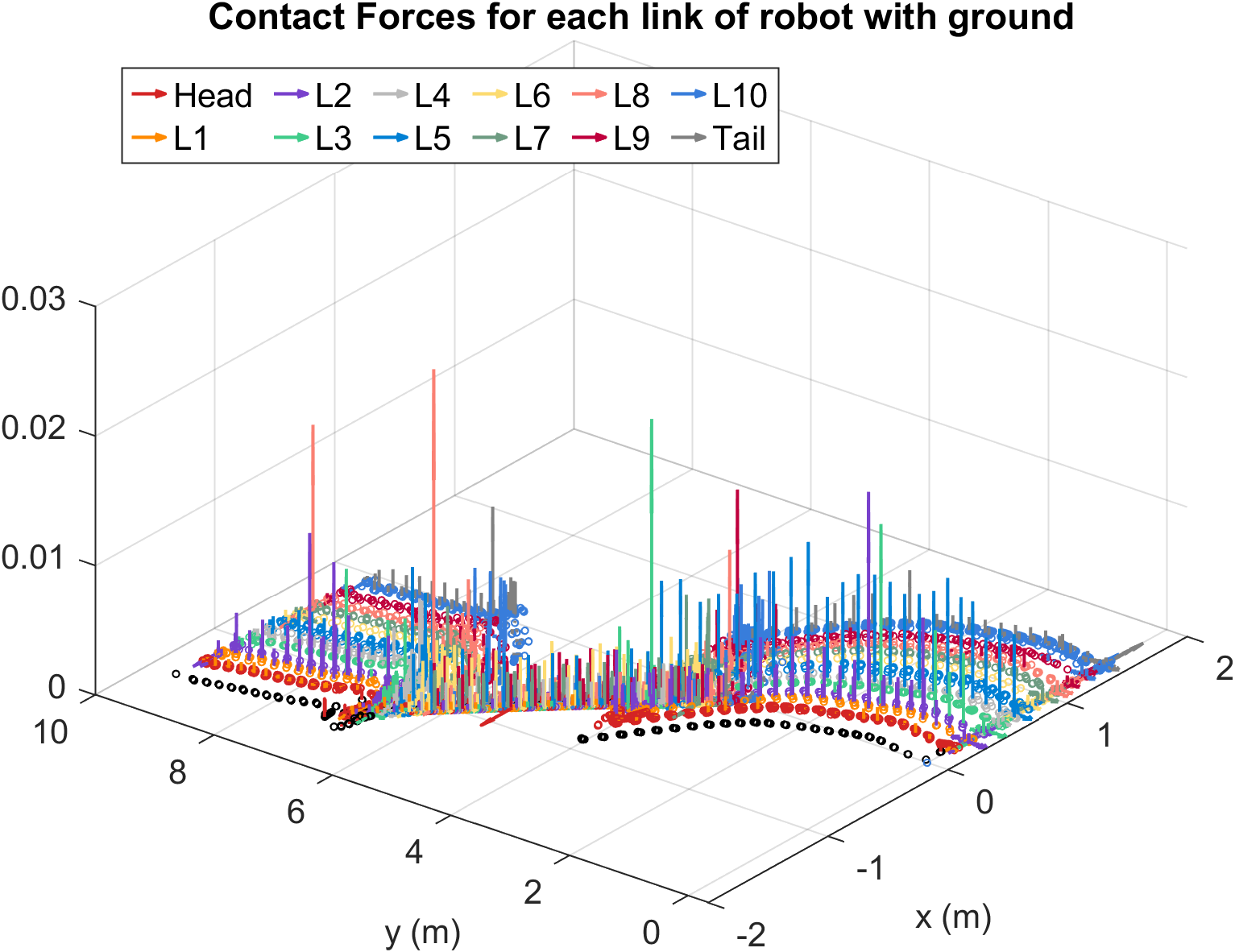} 
    \caption{Shows ground reaction forces for each link throught the locomotion in simulation}
    \label{fig:forces}
    \vspace{-5mm}
\end{figure}

Various components of the presented framework are validated in experiment and simulation. We have developed a high fidelity simulation of COBRA in MATLAB Simulink using the Simscape Multibody Toolbox. The accuracy of the simulation for predicting the movement of the real robot has been validated in a previous work \cite{salagame_reinforcement_2024}. Using this simulation we evaluate the presented reduced-order model and motion model. For a flat-ground CPG-based sidewinding trajectory of ten seconds, first we evaluate the ability of the ROM to accurately estimate link contacts according to equation \ref{eq:contact}. For an $\varepsilon$ of $0.015$, the estimated and actual contact signal for two links is shown in Figure \ref{fig:contact-signal}. This shows that the ROM is able to reasonably able to predict the link contacts, which is necessary for the motion model. 

We then evaluate the motion model by running prediction over the CPG trajectory at a rate of 200 Hz (the rate of operation of the CPG) for a horizon of 0.1 seconds, equivalent to 5\% of a gait cycle.  The prediction is reset at the end of each horizon and the prediction is started over. Figure \ref{fig:com-pred} shows the comparison of predicted CoM position vs actual over all horizons over the 10 second period. The top graph shows prediction in the XY plane with color indicating the passing of time. 

We then demonstrate the use of a CPG based trajectory to execute locomotion through a constrained path (Figure \ref{fig:experiment}) using the real robot. The robot, measuring 1.7m in length during its forward translation gait, uses the key actions identified in Figure \ref{fig:problem-statement} and travels through a corridor that is 0.3m wide. Here, the generated gaits are based on predefined CPG parameters, however this experiment serves as a proof of concept for implementation of an NMPC based controller for more complex environments.

To better study this locomotion and obtain data that is not available from hardware experiments, we mimic this setup in Simulink to perform narrow path locomotion using the same CPG generated gaits. This is shown in Figure \ref{fig:sim}. Using this, we can now study the distribution of ground contacts (Figure \ref{fig:contact}) and ground reaction forces (Figure \ref{fig:forces}) that is not feasible to measure in hardware. This setup will aid in validating the constraints applied in the NMPC framework and tuning the loss functions to achieve optimal performance.

\section{Concluding Remarks}
We propose an optimization-based motion planning methodology for snake robots operating in constrained environments. By leveraging a reduced-order modeling approach, we simplify the planning process, enabling the optimizer to effectively limit the robot’s footprint and autonomously generate gaits to reach target locations. Our approach is validated through high-fidelity simulations that accurately estimate contact interactions and validate the motion model. Key locomotion strategies were identified and successfully demonstrated in hardware experiments to traverse through a narrow corridor. This work advances the goal of fully autonomous gait generation and operation in confined spaces. Future efforts will focus on testing the NMPC framework both in simulation and hardware, alongside onboard continuous state estimation for full autonomy.
\balance{}
\printbibliography

\end{document}